\documentclass[10pt,twocolumn,letterpaper]{article}

\usepackage{cvpr}
\usepackage{times}
\usepackage{epsfig}
\usepackage{graphicx}
\usepackage{amsmath}
\usepackage{amssymb}
\usepackage{multirow}
\usepackage{array}
\usepackage{float}

\usepackage[breaklinks=true,bookmarks=false,colorlinks]{hyperref}

\cvprfinalcopy 

\def\cvprPaperID{****} 

\begin{document}

\title{Deep Learning for 3D Point Cloud Understanding: A Survey}

\author{Haoming Lu$^{1}$, Humphrey Shi$^{2,1}$
\\
\\
$^{1}$University of Illinois at Urbana-Champaign, $^{2}$University of Oregon\\
}

\maketitle

\begin{abstract}
   The development of practical applications, such as autonomous driving and robotics, has brought increasing attention to 3D point cloud understanding. 
   While deep learning has achieved remarkable success on image-based tasks, there are many unique challenges faced by deep neural networks in processing massive, unstructured and noisy 3D points. To demonstrate the latest progress of deep learning for 3D point cloud understanding, this paper summarizes recent remarkable research contributions in this area from several different directions (classification, segmentation, detection, tracking, flow estimation, registration, augmentation and completion), together with commonly used datasets, metrics and state-of-the-art performances. 
   More information regarding this survey can be found at: \href{https://github.com/SHI-Labs/3D-Point-Cloud-Learning}{https://github.com/SHI-Labs/3D-Point-Cloud-Learning}.
\end{abstract}

\section{Introduction}

\begin{figure}[t]
\begin{center}
\includegraphics[width=1.5in]{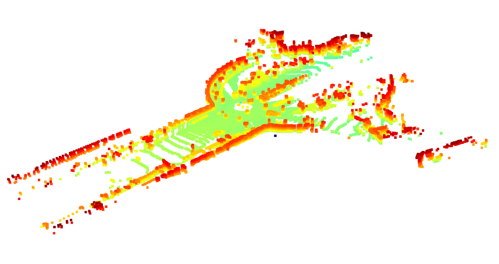}
\includegraphics[width=1.5in]{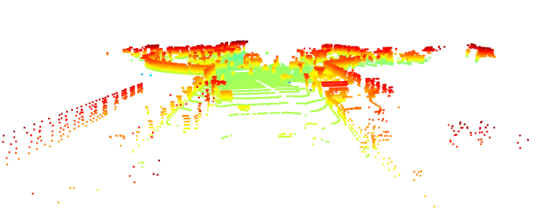}
\end{center}
   \caption{Point cloud data collected from outdoor scene, shown from two distinct angles.}
\label{fig:1}
\end{figure}

Deep learning has shown outstanding performance in a wide range of computer vision tasks in the past years, especially image tasks. Meanwhile, in many practical applications, such as autonomous vehicles (Figure \ref{fig:1} shows a point cloud collected by an autonomous vehicle), we need more information than only images to obtain a better sense of the environment. 3D data from lidar or RGB-D cameras are considered to be a good supplement here. These devices generate 3D geometric data in the form of point clouds. With the growing demand from industry, utilization of point clouds with deep learning models is becoming a research hotspot recently. \par

In constrast to image data, point clouds do not directly contain spatial structure, and deep models on point clouds must therefore solve three main problems: (1) how to find a representation of high information density from a sparse point cloud, (2) how to build a network satisfying necessary restrictions like size-variance and permutation-invariance, (3)  how to process large volumes of data with lower time and computing resource consumption. PointNet \cite{qi2017pointnet} is one of the representative early attempts to design a novel deep network for comsumption of unordered 3D point sets by taking advantage of MLP and T-Net. PointNet, together with its improved version PointNet++ \cite{qi2017pointnet++}, inspired a lot of follow-up works. \par
Fundamental tasks in images, such as classification, segmentation and object detection also exist in point clouds. Most solutions to these problems benefit from research findings on the image side, while adequate adaptions are inevitable to suit the characteristics of 3D data. In this paper, recent works on point clouds are divided into the following categories: classification, segmentation, detection, matching and registration, augmentation, completion and reconstruction. Detailed descriptions of each category will be provided in the following sections. \par
A growing number of datasets are available for different tasks on point clouds. ShapeNet \cite{shapenet2015} and ModelNet \cite{wu20153d} are two early datasets consisting of clean 3D models. These early datasets suffer from the lack of generalization. However, it is necessary to consider disturbance including noise and missing points to develop robust models. With that in mind, datasets such as ScanNet \cite{dai2017scannet} and KITTI \cite{Geiger2013IJRR} are then created from scans of the actual environment. Datasets designed for autonomous vehicle tasks, like nuScenes \cite{caesar2019nuscenes} and Lyft \cite{lyft2019}, are further generalized by involving various environments at different times. Currently, ever more datasets are being proposed in order to meet the increasing demands of distinct niches. \par

The structure of this paper is as follows. Section 2 introduces existing 3D datasets and corresponding metrics for different tasks. Section 3 includes a survey of 3D shape classification methods. Section 4 reviews methods for 3D semantic segmentation and instance segmentation. Section 5 presents a survey of methods for 3D object detection and its derivative task. Section 6 introduces recent progress in 3D point cloud matching and registration. Section 7 provides a review of methods to improve data quality. Finally, section 8 concludes the paper. \par

\section{Datasets and metrics}
Datasets are of great importance in deep learning methods for 3D point cloud data. First, well-designed datasets provide convictive evaluation and comparison among different algorithms. Second, datasets with richer content and metadata help define more complicated tasks and raise new research topics. In this section, we will briefly introduce some most commonly used datasets and evaluation metrics.

\begin{table*}[h]
\centering
\caption{Commonly used 3D point cloud datasets in recent works}
\begin{tabular}{|p{2.2cm}|p{2cm}|p{1cm}|p{2.8cm}|p{5.5cm}|p{0.7cm}|}
\hline
Dataset & Task & Classes & Scale & Feature & Year \\
\hline
ShapeNet \cite{shapenet2015} & Classification & 55 & 51300 models & The categories are selected according to WordNet \cite{miller1995wordnet} synset. & 2015\\
\hline
ModelNet40 \cite{wu20153d} & Classification & 40 & 12311 models & The models are collected with online search engines by querying for each established object category. & 2015 \\
\hline
S3DIS \cite{armeni20163d} & Segmentation & 12 & 215 million points & Points are collected in 5 large-scale indoor scenes from 3 different buildings. & 2016 \\
\hline
Semantic3D \cite{hackel2017isprs} & Segmentation & 8 & 4 billion points & Hand-labelled from a range of diverse urban scenes. & 2017 \\
\hline
ScanNet \cite{dai2017scannet} & Segmentation & 20 & 2.5 million frames & Collected with a scalable RGB-D capture system with automated surface reconstruction and crowdsourced semantic annotation. & 2017 \\
\hline
KITTI \cite{Geiger2013IJRR,Geiger2012CVPR,Fritsch2013ITSC,Menze2015CVPR} & Detection Tracking & 3 & 80256 objects & Captured by a standard station wagon equipped with two cameras, a Velodyne laser scanner and a GPS localization system driving in different outdoor scenes. & 2012\\
\hline
nuScenes \cite{caesar2019nuscenes} & Detection Tracking & 23 & 1.4M objects & Captured with full sensor suite (1x LIDAR, 5x RADAR, 6x camera, IMU, GPS); 1000 scenes of 20s each. & 2019\\
\hline
Waymo Open Dataset \cite{sun2020scalability} & Detection Tracking & 4 & 12.6M objects with tracking ID & Captured with 1 mid-range lidar, 4 short-range lidars
and 5 cameras (front and sides); 1,950 segments of 20s each, collected at 10Hz. & 2019 \\
\hline
\end{tabular}
\label{table:0}
\end{table*}

\subsection{Datasets}
Table \ref{table:0} shows the most commonly used 3D point cloud datasets for three matured tasks (classification, segmentation and detection), which will be mentioned often in the following sections. We will also introduce each of them with more details. \par 
\textbf{ShapeNet} ShapeNet \cite{shapenet2015} is a rich-annotated dataset with 51300 3D models in 55 categories. It consists of several subsets. ShapeNetSem, which is one of the subsets, contains 12000 models spread over a broader set of 270 categories. This dataset, together with ModelNet40 \cite{wu20153d}, are relatively clean and small, so they are usually used to evaluate the capacity of backbones before applied to more complicated
tasks.\par
\textbf{ModelNet40} The ModelNet \cite{wu20153d} project provides three benchmarks: ModelNet10, ModelNet40 and Aligned40. The ModelNet40 benchmark, where ``40" indicates the number of classes, is the most widely used. To find the most common object categories, the statistics obtained from the SUN database \cite{xiao2010sun} are utilized. After establishing the vocabulary, 3D CAD models are collected with online search engines and verified by human workers. \par
\textbf{S3DIS} The Stanford Large-Scale 3D Indoor Spaces (S3DIS) dataset is composed of 5 large-scale indoor scenes from three buildings to hold diverse in architectural style and appearance. The point clouds are automatically generated without manual intervention. 12 semantic elements including structural elements (floor, wall, etc.) and common furniture are detected. \par
\textbf{Semantic3D} Semantic3D \cite{hackel2017isprs} is the largest 3D point cloud dataset for outdoor scene segmentation so far. It contains over 4 billion points collected from around 110000$m^2$ area with a static lidar. The natural of outdoor scene, such as the unevenly distribution of points and massive occlusions, makes the dataset challenging. \par
\textbf{ScanNet} ScanNet \cite{dai2017scannet} is a video dataset consists of 2.5 million frames from more than 1000 scans, annotated with camera poses, surface reconstructions and instance-level semantic segmentation. The dataset provides benchmarks for mutiple 3D scene understanding tasks, such as classification, semantic voxel labeling and CAD model retrieval. \par
\textbf{KITTI} The KITTI \cite{Geiger2013IJRR,Geiger2012CVPR,Fritsch2013ITSC,Menze2015CVPR} vision benchmark suite is among the most famous benchmarks with 3D data. It covers benchmarks for 3D object detection, tracking and scene flow estimation. The multi-view data are captured with an autonomous driving platform with two high-resolution color and gray cameras, a Velodyne laser scanner and a GPS localization system. Only three 
kinds of objects which are important to autonomous driving are labelled: cars, pedestrians and cyclists.\par
\textbf{Other datasets} There are some other datasets of high quality but not widely used, such as Oakland \cite{munoz2009contextual}, iQmulus \cite{vallet2015terramobilita} and Paris-Lille-3D \cite{roynard2017parisIJRR}. 3DMatch \cite{zeng20183dcontextnet} pushed the research in 3D matching and registration, which is a less popular direction in the past period. Recently, the rising demand from industry of autonomous driving has spawned several large-scale road-based datasets, represented by nuScenes \cite{caesar2019nuscenes}, Lyft Level 5 \cite{lyft2019} and Waymo Open Dataset \cite{sun2020scalability}. They proposed complicated challenges requiring to leverage multi-view data and related metadata. The development of datasets is helping reduce the gap between research and practical applications.
\subsection{Metrics}
The comparison between different algorithms requires certain metrics. It is important to design and select appropriate metrics. Well-designed metrics can provide valid evaluation of different models, while unreasonable metrics might lead to incorrect conclusions. \par
Table \ref{table:metrics} lists widely used metrics in different tasks. For classification methods, overall accuracy and mean accuracy are most frequently used. Segmentation models can be analyzed by accuracy or (m)IoU. In detection tasks, the result are usually evaluated region-wise, so (m)IoU, accuracy, precision and recall could apply. MOTA and MOTP are specially designed for object tracking modelts, while EPE is for scene for estimation. ROC curves, which is the derivative of precision and recall, help evaluate the performance of 3D match and registration models. Besides, visualization is always an effective supplement of numbers.
\begin{table*}[htbp]
\centering
\caption{Commonly used metrics for different tasks. In this table, $N$ denotes the number of samples, $C$ denotes the number of categories, $IDS$ denotes the number of identity switches, $I_{i,j}$ denotes the number of points that are from ground truth class/instance $i$ and labelled as $j$, $TP/TN/FP/FN$ stands for the number of true positives, true negatives, false positives and false negatives respectively. Higher metrics indicate better results if not specified otherwise.}
\begin{tabular}{|c|c|c|c|}
\hline
Metric & Formula & Explanation \\
\hline
Accuracy & $Accuracy=\frac{TP+TN}{TP+TN+FP+FN}$ & \multicolumn{1}{|m{9cm}|}{Accuracy indicates how many predictions are correct over all predictions. ``Overall accuracy (OA)" indicates the accuracy on the entire dataset.} \\
\hline
mACC & $mACC=\frac{1}{C}\sum_{c=1}^C Accuracy_c$ & \multicolumn{1}{|m{9cm}|}{The mean of accuracy on different categories, useful when the categories are imbalanced.} \\
\hline
Precision & $Precision=\frac{TP}{TP+FP}$ & \multicolumn{1}{|m{9cm}|}{The ratio of correct predictions over all predictions.} \\
\hline
Recall & $Recall=\frac{TP}{TP+FN}$ & \multicolumn{1}{|m{9cm}|}{The ratio of correct predictions over positive samples in the ground truth.} \\
\hline
F1-Score & $F_1=2\times \frac{Precision\cdot Recall}{Precision + Recall}$ & \multicolumn{1}{|m{9cm}|}{The harmonic mean of precision and recall.} \\
\hline
IoU & $IoU_i=\frac{I_{i,i}}{\sum_{c=1}^C (I_{i,c}+I_{c,i}) - I_{i,i}}$ & \multicolumn{1}{|m{9cm}|}{Intersection over Union (of class/instance $i$). The intersection and union are calculated between the prediction and the ground truth.}\\
\hline
mIoU & $mIoU = \frac{1}{C}\sum_{c=1}^{C}IoU_i$ & \multicolumn{1}{|m{9cm}|}{The mean of IoU on all classes/instances.} \\
\hline
MOTA & $MOTA=1-\frac{FN+FP+IDS}{TP+FN}$ & \multicolumn{1}{|m{9cm}|}{Multi-object tracking accuracy (MOTA) synthesizes 3 error sources: false positives, missed targets and identity switches, and the number of ground truth (as $TP+FN$) is used for normalization.} \\
\hline
MOTP & $MOTP=\frac{\sum_{i,t} e_{i,t}}{\sum_t d_t}$ & \multicolumn{1}{|m{9cm}|}{Multi-object tracking precision (MOTP) indicates the precision of localization. $d_t$ denotes the number of matches at time $t$, and $e_{i,t}$ denotes the error of the $i$-th pair at time $t$.} \\
\hline
EPE & $EPE=||\Hat{sf}-sf||_2$ & \multicolumn{1}{|m{9cm}|}{End point error (EPE) is used in scene flow estimation, also referred as EPE2D/EPE3D for 2D/3D data respectively. $\Hat{sf}$ denotes the predicted scene flow vector while $sf$ denotes the ground truth.} \\ 
\hline
\end{tabular}
\label{table:metrics}
\end{table*}

\section{Classification}
\subsection{Overview}
\begin{figure*}[h]
\begin{center}
\includegraphics[width=5in]{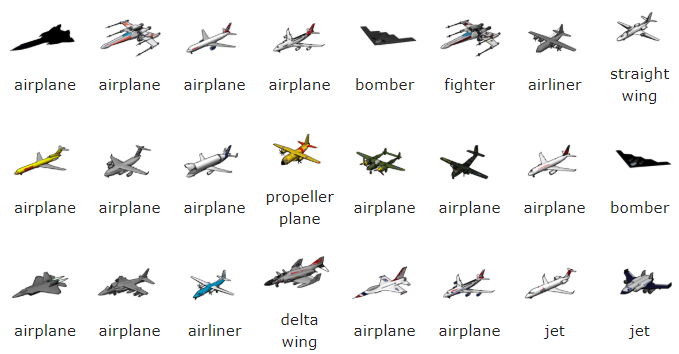}
\end{center}
   \caption{3D models from ShapeNet \cite{shapenet2015}. ShapeNet contains large-scale 3D models with manually verified annotation.}
\label{fig:2}
\end{figure*}

Classification on point clouds is commonly known as 3D shape classification. Similar to image classification models, models on 3D shape classification usually first generate a global embedding with an aggregation encoder, then pass the embedding through several fully connected layers to obtain the final result. Most 3D shape classification methods are tested with clean 3D models (as in Figure \ref{fig:2}). Based on the point cloud aggregation method, classification models can be generally divided into two categories: projection-based methods and point-based methods.

\subsection{Projection-based Methods}
Projection-based methods project the unstructured 3D point clouds into specific presupposed modality (e.g. voxels, pillars), and extract features from the target format, which allows them to benefit from the previous research findings in the corresponding direction. \par

\subsubsection{Multi-view representation}
MvCNN \cite{su15mvcnn} is a method based on a multi-view representation of point clouds. A 3D point cloud is represented by a group of 2D images by rendering snapshots from different angles. Each image in the group will be passed through a CNN to extract view-based features, pooled across views and passed through another CNN to build a compact descriptor. While MVCNN does not distinguish different views, it is helpful to consider the relationship among views. GVCNN \cite{feng2018gvcnn} is a method that takes advantage of this relationship. By quantifying the discrimination of views, we are able to divided the set of views into groups based on their discrimination scores. The view descriptors will be passed through intra-group pooling and cross-group fusion for prediction. Aside from the models mentioned above, \cite{yu2018multi} and \cite{yang2019learning} also improve the recognition accuracy with multi-view representation.

\subsubsection{Volumetric representation}
VoxNet \cite{maturana2015voxnet} is an early method using the volumetric representation. In this method, each point $(x, y, z)$ is projected into a corresponding discrete voxel point $(i, j, k)$. Each point cloud will be mapped into an occupancy grid of $32\times 32\times 32$ voxels, and the grid will then be passed through two 3D convolutional layers to obtain the final representation. \par
VoxNet simply uses adaption of CNN layers for the prediction head, which leads to potential loss of detailed spatial information. 3D ShapeNet \cite{wu20153d} proposed a belief-based deep convolutional network to learn the distribution of point clouds in different 3D shapes. In this method, 3D shapes are represented by the probability distributions of binary variables on grids. \par
While volumetric methods already achieve satisfactory performance, most suffer from the cubic growth of computation complexity and memory footprint, hence the resolution of the grid is strictly limited. OctNet \cite{riegler2017octnet} improved the efficiency by introducing a hybrid grid-octree structure to hierarchically partition point clouds. A point cloud is represented by several octrees along a regular grid, each octree is encoded as a bit string, and features are generated through naive arithmetic. Inspired by OctNet, OCNN \cite{wang2017cnn} then proposed a method that introduces 3D-CNNs to extract features from octrees. \par
Methods based on volumetric representations as mentioned above are naturally coarse as only a small fraction of voxels are non-empty and the detailed context inside each voxel is hardly collected. The balance between resolution and computation is difficult to achieve in practice. \par

\subsubsection{Basis point set}
BPS \cite{prokudin2019efficient} proposed a new approach that breaks the convention that point clouds, even with various sizes, are usually projected onto a grid of same size. In BPS, input points are first normalized into a unit ball, then a group of points is randomly sampled to make up a basis point set (BPS). The sampled BPS is constant for all point clouds in a dataset. For a given point cloud $X$, each point $x_i$ is represented by the Euclidean distance between itself and its nearest neighbor in BPS. By passing such representation through the last two fully connected layers of PointNet, the model achieves performance similar to that of the original PointNet design.

\subsection{Point-based Methods}
Compared with projection-based methods that aggregate points from a spatial neighborhood, point-based methods attempt to learn features from individual points. Most of recent work focuses on this direction.

\subsubsection{MLP networks}
PointNet \cite{qi2017pointnet} is a famous architecture that takes advantage of multi-layer perceptrons (MLPs). The input (an $n \times 3$ 2D tensor) is first multiplied by an affine transformation matrix
predicted by a mini-network (T-Net) to hold invariance under geometric transformations. The point set is then passed through a group of MLPs followed by another joint alignment network, and a max-pooling layer to obtain the final global feature. This backbone can be used for both classification and segmentation prediction. For classification, the global feature is passed through an MLP for output scores. For segmentation, the concatenations of the global feature and different levels of intermediate features from each point are passed through an MLP for the classification result of each point. Conventional CNNs take features at different scales by a stack of convolutional layers; inspired by that, PointNet++ \cite{qi2017pointnet++} is proposed. In this work, the local region of a point $x$ is defined as the points within a sphere centered at $x$. One set abstraction level here contains a sampling layer, a grouping layer to identify local regions and a PointNet layer. Stacking such set abstraction levels allows us to extract features hierarchically as CNNs for image tasks do. \par
The simple implementation and promising performance of PointNet \cite{qi2017pointnet} and PointNet++ \cite{qi2017pointnet++} inspired a lot of follow-up work. PointWeb \cite{zhao2019pointweb} is adapted from PointNet++ and improves quality of features by introducing Adaptive Feature Adjustment (AFA) to make use of context information of local neighborhoods. In addition, SRN \cite{duan2019structural} proposed Structural Relation Network (SRN) to equip PointNet++, and obtained better performance.

\subsubsection{Convolutional networks}
Convolution kernels on 2D data can be extended to work on 3D point cloud data. As mentioned before, VoxNet \cite{maturana2015voxnet} is an early work that directly takes advantage of 3D convolution. \par
A-CNN \cite{komarichev2019cnn} proposed another way to apply convolution on point clouds. In order to prevent redundant information from overlapped local regions (the same group of neighboring points might be repeatedly included in regions at different scales), A-CNN proposed a ring-based scheme instead of spheres. To convolve points within a ring, points are projected on a tangent plane at a query point $q_i$, then ordered in clockwise or counter-clockwise direction by making use of cross product and dot product, and eventually a 1-D convolution kernel will be applied to the ordered sequence. The output feature can be used for both classification and segmentation as in PointNet.\par
RS-CNN \cite{liu2019relation} is another convolutional network based on relation-shape convolution. An RS-Conv kernel takes a neighborhood around a certain point as its input, and learns the mapping from naive relations (e.g. Euclidean distance, relative position) to high-level relations among points, and encodes the spatial structure within the neighborhood with the learned mapping.\par
In PointConv \cite{wu2019pointconv}, the convolution operation is defined as finding a Monte Carlo estimation of the hidden continuous 3D convolution w.r.t. an importance sampling. The process is composed with a weighting function and a density function, implemented by MLP layers and a kernelized density estimation. Furthermore, the 3D convolution is reduced into matrix multiplication and 2D convolution for memory and computational efficiency and easy deployment. A similar idea is used in MCCNN \cite{hermosilla2018monte}, where convolution is replaced by a Monte Carlo estimation based on the density function of the sample. \par
Geo-CNN \cite{lan2019modeling} proposed another way to model the geometric relationship among neighborhood points. By taking six orthogonal bases, the space will be separated into eight quadrants, and all vectors in a specific quadrant can be composed by three of the bases. Features are extracted independently along each direction with corresponding direction-associated weight matrices, and are aggregated based on the angle between the geometric vector and the bases. The feature of some specific point at the current layer is the sum of features of the given point and its neighboring edge features from the previous layer. \par
In SFCNN \cite{rao2019spherical}, the input point cloud is projected onto regular icosahedral lattices with discrete sphere coordinates, hence convolution can be implemented by maxpooling and convolution on the concatenated features from vertices of spherical lattices and their neighbors. SFCNN holds rotation invariance and is robust to perturbations.

\subsubsection{Graph networks}
Graph networks consider a point cloud as a graph and the vertices of the graph as the points, and edges are generated based on the neighbors of each point. Features will be learned in spatial or spectral domains. \par
ECC \cite{simonovsky2017dynamic} first proposed the idea of considering each point as a vertex of the graph and connected edges between pairs of points that are ``neighbors". Then, edge conditioned convolution (ECC) is applied with a filter generating network such as MLP. Neighborhood information is aggregated by maxpooling and coarsened graph will be generated with VoxelGrid \cite{rusu20113d} algorithm. After that, DGCNN \cite{wang2019dynamic} uses a MLP to implement EdgeConv, followed by channel-wise symmetric aggregation on edge features from the neighborhood of each point, which allows the graph to be dynamically updated after each layer of the network. \par
Inspired by DGCNN, Hassani and Haley \cite{hassani2019unsupervised} proposed an unsupervised multi-task approach to learn shape features. The approach consists of an encoder and an decoder, where the encoder is constructed from multi-scale graphs, and the decoder is constructed for three unsupervised tasks (clustering, self-supervised classification and reconstruction) trained by a joint loss. \par
ClusterNet \cite{chen2019clusternet} uses rigorously rotation-invariant (RRI) module to generate rotation-invariant features from each point, and an unsupervised agglomerative hierarchical clustering method to construct hierarchical structures of a point cloud. Features of sub-clusters at each level are first learned with an EdgeConv block, then aggregated by maxpooling. \par

\subsubsection{Other networks}
Aside from OctNet \cite{riegler2017octnet}, which uses octrees on voxel grids to hierarchically extract features from point clouds, Kd-Net \cite{klokov2017escape} makes use of K-d trees to build a bottom-up encoder. Leaf node representations are normalized 3D coordinates (by setting the center of mass as origin and rescaled to $[-1,1]^3$), and non-leaf node representations are calculated from its children nodes with MLP. The parameters of MLPs are shared within each level of the tree. Moreover, 3DContextNet \cite{zeng20183dcontextnet} proposed another method based on K-d trees. While non-leaf representations are still computed with MLP from its children, the aggregation at each level is more complicated for considering both local cues and global cues. The local cues concern points in the corresponding local region, and the global cues concern the relationship between current position and all positions in the input feature map. The representation at the root will be used for prediction. \par
RCNet \cite{wu2019point} introduced RNN to point cloud embedding. The ambient space is first partitioned into parallel beams, each beam is then fed into a shared RNN, and the output subregional features are considered as a 2D feature map and processed by a 2D CNN. \par
SO-Net \cite{li2018so} is a method based on the self-organized map (SOM). A SOM is a low-dimensional (two-dimensional in the paper) representation of the input point cloud, initialized by a proper guess (dispersing nodes uniformly in a unit ball), and trained with unsupervised competitive learning. A k-nearest-neighbor set is searched over the SOM for each point, and the normalized KNN set is then passed through a series of fully connected layers to generate individual point features. The point features are used to generate node features by maxpooling according to the association in KNN search, and the node features are passed through another series of fully connected layers and aggregated into a global representation of the input point cloud.

\subsection{Experiments}
Different methods choose to test their models on various datasets. In order to obtain a better comparison among methods, we select datasets that most methods are tested on, and list the experiment results for them in Table \ref{table:1}.

\begin{table*}[htbp]
\centering
\caption{Experiment results on ModelNet40 classification benchmark. ``OA" stands for overall accuracy and ``mACC" stands for mean accuracy.}
\begin{tabular}{|c|c|c|}
\hline
Methods & ModelNet40(OA) & ModelNet40(mAcc) \\
\hline
PointNet \cite{qi2017pointnet} & 89.2\% & 86.2\% \\
\hline
PointNet++ \cite{qi2017pointnet++} & 90.7\% & 90.7\% \\
\hline
PointWeb \cite{zhao2019pointweb} & 92.3\% & 89.4\% \\
\hline
SRN \cite{duan2019structural} & 91.5\% & - \\
\hline
Pointwise-CNN \cite{hua2018pointwise} & 86.1\% & 81.4\% \\
\hline
PointConv \cite{wu2019pointconv} & 92.5\% & - \\
\hline
RS-CNN \cite{liu2019relation} & 92.6\% & - \\
\hline
GeoCNN \cite{lan2019modeling} & 93.4\% & 91.1\% \\
\hline
A-CNN \cite{komarichev2019cnn} & 92.6\% & 90.3\% \\
\hline
Hassani and Haley \cite{hassani2019unsupervised} & 89.1\% & - \\
\hline
ECC \cite{simonovsky2017dynamic} & 87.4\% & 83.2\% \\
\hline
SFCNN \cite{rao2019spherical} & 91.4\% & - \\
\hline
DGCNN \cite{wang2019dynamic} & 92.2\% & 90.2\% \\
\hline
ClusterNet \cite{chen2019clusternet} & 87.1\% & - \\
\hline
BPS \cite{prokudin2019efficient} & 91.6\% & - \\
\hline
KD-Net \cite{klokov2017escape} & 91.8\% & 88.5\% \\
\hline
3DContextNet\cite{zeng20183dcontextnet} & 91.1\% & - \\
\hline
RCNet \cite{wu2019point} & 91.6\% & - \\
\hline
SO-Net \cite{li2018so} & 90.9\% & 87.3\% \\
\hline
\end{tabular}
\label{table:1}
\end{table*}

\section{Segmentation}

\subsection{Overview}
\begin{figure}[htbp]
\begin{center}
\includegraphics[width=3.2in]{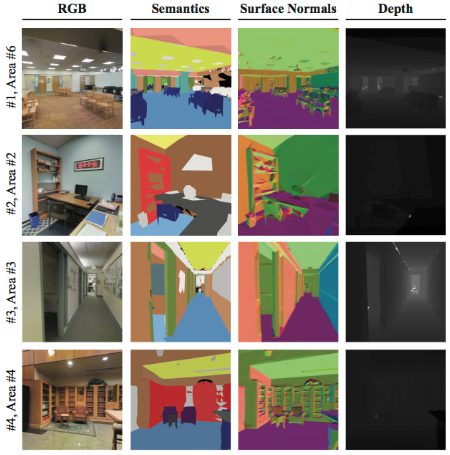}
\end{center}
   \caption{Stanford Large-Scale 3D Indoor Spaces Dataset \cite{armeni20163d} (S3DIS).}
\label{fig:3}
\end{figure}

3D segmentation intends to label each individual point, which requires the model to collect both global context and detailed local information at each point. Figure \ref{fig:3} shows some examples from S3DIS \cite{armeni20163d} dataset. There are two main tasks in 3D segmentation: semantic segmentation and instance segmentation.\par
Since a large number of classification models are able to achieve very high performance on popular benchmarks, they tend to test their backbone on segmentation datasets to prove the novel contribution and generalization ability. We will not reintroduce these models if they have been mentioned above. There are also some models that benefit from the jointly training on multiple tasks, and we will discuss these methods later in section 3.4.

\subsection{Semantic Segmentation}
Similar to 3D shape classification models, based on how the raw point cloud is organized, semantic segmentation methods can be generally divided into projection-based methods and point-based methods.

\subsubsection{Projection-based methods}
Huang and You \cite{huang2016point} project the input point cloud into occupancy voxels, which are then fed into a 3D convolutional network to generate voxel-level labels. All points within a voxel are assigned with the same semantic label as the voxel. ScanComplete \cite{dai2018scancomplete} utilizes fully convolutional networks to adapt to different input data sizes, and deploys a coarse-to-fine strategy to improve the resolution of predictions hierarchically. VV-Net \cite{meng2019vv} also transfers unordered points into regular voxel grids as the first step. After that, the local geometry information of each voxel will be encoded with a kernel-based interpolated variational auto-encoder (VAE). In each voxel, a radial basis function (RBF) is computed to generate a local continuous representation to deal with sparse distributions of points.  \par
F. Jaremo-Lawin et al. \cite{lawin2017deep} proposed a multi-view method that first projects a 3D cloud to 2D planes from multiple camera views, then pixel-wise scores on synthetic images are predicted with a multi-stream FCN, and the final labels are obtained by fusing scores over different views. PolarNet \cite{Zhang_polar_2020_CVPR}, however, proposed a polar BEV representation. By implicitly aligning attention with the long-tailed distribution, this representation reduces the imbalance of points across grid cells along the radial axis.\par 
Some other methods leverage scans in multiple modalities. 3DMV \cite{dai20183dmv} proposed a joint 3D-multi-view network that combines features from RGB images and point cloud. Features are extracted with a 3D CNN stream and a group of 2D streams respectively. MVPNet \cite{jaritz2019multi} proposed another aggregation to fuse features (from images and point cloud) in 3D canonical space with a point-based network. \par
\par

\subsubsection{Point-based methods}
First of all, PointNet \cite{qi2017pointnet} and PointNet++ \cite{qi2017pointnet++} can predict semantic labels with corresponding prediction branches attached. Engelmann et al. \cite{engelmann2018know} proposed a method to define neighborhoods in both world space and feature space with k-means clustering and KNN. A pairwise distance loss and centroid loss are introduced to feature learning based on the assumption that points with the same semantic label are supposed to be closer. PointWeb \cite{zhao2019pointweb}, as mentioned in classification, can also be adapted to predict segmentation labels. PVCNN \cite{liu2019point} proposed a comprehensive method that leverages both point and voxel representation to obtain memory and computation efficiency simultaneously.\par
Some extensions of the convolution operator are introduced for feature extraction on point cloud. PCCN \cite{wang2018deep} introduces parametric continuous convolutional layers. These layers are parameterized by MLPs and span full continuous vector spaces. The generalization allows models to learn over any data structure where the support relationship is computable. Pointwise-CNN \cite{hua2018pointwise} introduced a point-wise convolution where the neighbor points are projected into kernel cells and convolved with corresponding kernel weights. Engelmann et al. \cite{engelmann2019dilated} proposed Dilated Point Convolution (DPC) to aggregate dilated neighbor features, instead of the conventional k-nearest neighbors. \par
Graph networks are also used in some segmentation models to obtain the underlying geometric structures of the input point clouds. SPG \cite{landrieu2018large} introduced a structure called superpoint graph (SPG) to capture the organization of point clouds. The idea is further extended in \cite{landrieu2019point}, which introduces a oversegmentation (into pure superpoints) of the input point cloud. Aside from that, Graph Attention Convolution \cite{wang2019graph} (GAC) is proposed to learn relevant features from local neighborhoods selectively. By dynamically assigning attention weights to different neighbor points and different feature channels based on their spatial positions and feature differences, the model is able to learn discriminative features from the most relevant part of the neighbor point sets.\par
Compared with projection-based methods, point-based methods usually require more computation and therefore have more trouble dealing with large-scale data. Tatarchenko et al. \cite{tatarchenko2018tangent} introduced tangent convolutions to solve this. A fully-convolutional network is designed based on the tangent convolution and successfully improved the performance on large-scale point clouds. RandLA-Net \cite{Hu_rand_2020_CVPR} attempted to reduce computation by replace conventional complex point sampling approaches with random sampling. And to avoid random sampling from discarding crucial information, a novel feature aggregation module is introduced to enlarge receptive fields of each point. \par
Based on the fact that the production of point-level labels is labor-intensive and time-consuming, some methods explored weakly supervised segmentation. Xu and Lee \cite{xulee2020weakly} proposed a weakly supervised approach which only requires a small fraction of points to be labelled at training stage. By learning gradient approximation and smoothness constraints in geometry and color, competitive results can be obtained with as few as 10\% points labelled. On the other hand, Wei et al. \cite{Wei_multi_2020_CVPR} introduced a multi-path region mining module, which can provide pseudo point-level labels by a classification network over weak labels. The segmentation network is then trained with these pseudo labels in a fully supervised manner. 

\subsection{Instance Segmentation}
Instance segmentation, compared with semantic segmentation, requires distinguishing points with same semantic meaning, which makes the task more challenging. In this section, instance segmentation methods are further divided into two categories: proposal-based methods and proposal-free methods.

\subsubsection{Proposal-based methods}
Proposal-based instance segmentation methods can be considered as the combination of object detection and mask prediction. 3D-SIS \cite{hou20193d} is a fully convolutional network for 3D semantic instance segmentation where geometry and color signals are fused. For each image, 2D features for each pixel are extracted by a series of 2D convolutional layers, and then backprojected to the associated 3D voxel grids. The geometry and color features are passed through a series of 3D convolutional layers respectively and concatenated into a global semantic feature map. Then a 3D-RPN and a 3D-RoI layer are applied to generate bounding boxes, instance masks and object labels. Generative Shape Proposal Network (GSPN) \cite{yi2019gspn} generates proposals by reconstructing shapes from the scene instead of directly regresses bounding boxes. The generated proposals are refined with a region-based PointNet (R-PointNet), and the labels are determined with a point-wise binary mask prediction over all class labels. 3D-BoNet \cite{yang2019learning} is a single-stage method that adapts PointNet++ \cite{qi2017pointnet++} as backbone network to global features and local features at each point. Two prediction branches follow to generate instance-level bounding box and point-level mask respectively. Zhang el al. \cite{zhang2020instance} proposed a method for large-scale outdoor point clouds. The point cloud is first encoded into a high-resolution BEV representation augmented by KNN, and features are then extracted by voxel feature encoding (VFE) layers and self-attention blocks. For each grid, a horizontal object center and its height limit are predicted, objects that are closed enough will be merged, and eventually these constraints will be leveraged to generate instance prediction.

\subsubsection{Proposal-free methods}
Proposal-free methods tend to generate instance-level label based on semantic segmentation by algorithms like clustering. Similarity Group Proposal Network (SGPN) \cite{wang2018sgpn} is a representative work that learns a feature and semantic map for each point, and a similarity matrix to estimate the similarity between pairs of features. A heuristic non-maximal suppression method follows to merge points into instances. Lahoud et al. \cite{lahoud20193d} adopted multi-task metric learning to (1) learn a feature embedding such that voxels with the same instance label are close and those with different labels are separated in the feature space and (2) predict the shape of instance at each voxel. Instance boundaries are estimated with mean-shift clustering and NMS. \par
Zhang et al. \cite{zhang2019point} introduced a probabilistic embedding to encode point clouds. The embedding is implemented with multivariate
Gaussian distribution, and the Bhattacharyya kernel is adopted to esimate the similarity between points. Proposal-free methods do not suffer from the computational complexity of region-proposal layers; however, it is usually difficult for them to produce discriminative object boundaries from clustering. \par
There are also several instance segmentation methods based on projection. SqueezeSeg \cite{wu2018squeezeseg} is one of the pioneer works in this direction. In this method, points are first projected onto a sphere for a grid-based representation. The transformed representation is of size $H\times W\times C$, where in practice $H$=64 is the number of vertical channels of lidar, $W$ is manually picked to be 512, and $C$ equals to 5 (3 dimensional coordinates + intensity measurement + range). The representation is then fed through a conventional 2D CNN and a conditional random field (CRF) for refined segmentation results. This method is afterwards improved by SqueezeSegv2 \cite{wu2019squeezesegv2} with a context aggregation module and a domain adaptation pipeline. \par
The idea of projection-based methods is further explored by Lyu et al. \cite{Lyu_Seg2D_2020_CVPR}. Inspired by graph drawing algorithms, they proposed a hierarchical approximate algorithm to project point clouds into image representations with abundant local geometric information preserved. The segmentation will then be generated by a multi-scale U-Net from the image representation. With this innovative projection algorithm, the method obtained significant improvement. \par
PointGroup \cite{Jiang_pointgroup_2020_CVPR} proposed a bottom-up framework with two prediction branches. For each point, its semantic label and relative offset to its respective instance centroid are predicted. The offset branch helps better grouping of points into objects as well as separation of objects with the same semantic label. During the clustering stage, both original positions and shifted positions are considered, the association of these two results turns out to have a better performance. Along with NMS based on the newly designed ScoreNet, this method outperforms other works of the day by a great margin. \par

\subsection{Joint Training}
As mentioned above, some recent works jointly address more than one problems to better realized the power of models. The unsupervised multi-task approach proposed by Hassani and Haley \cite{hassani2019unsupervised} is an example in which clustering, self-supervised classification and reconstruction are jointly trained. The two tasks under segmentation, semantic segmentation and instance segmentation, are also proven to likely benefit from simultaneous training. \par
There are two naive ways to solve semantic segmentation and instance segmentation at the same time: (1) solve semantic segmentation first, run instance segmentation on points of certain labels based on the result of semantic segmentation, (2) solve instance segmentation first, and directly assign semantic labels with instance labels. These two step-wise paradigms highly depend on the output quality of the first step, and are not able to make full use of the shared information between two tasks. \par 
JSIS3D \cite{pham2019jsis3d} develops a pointwise network that predicts the semantic label of each point and high-dimensional embeddings at the same time. After these steps, instances of the same class will have similar embeddings, then a multi-value conditional random field model is applied to synthesize semantic and instance labels, formulating the problem as jointly optimizing labels in the field model. ASIS \cite{wang2019associatively} is another method that makes the two tasks benefit from each other. Specifically, instance segmentation benefits from semantic segmentation by learning semantic-aware instance embedding at point level, while semantic features of the point set from the same instance will be fused together to generate accurate semantic predictions for every point.

\subsection{Experiments}
We select the benchmarks on which most methods are tested, S3DIS\cite{armeni20163d}, to compare the performance of different methods. The performances are summarized in Table \ref{table:2}.

\begin{table*}[!htbp]
\centering
\caption{Experiment results on on semantic segmentation in S3DIS benchmark. Only results that are reported in the original papers are listed, those reported as a reference by other papers are excluded because they are sometimes conflicting.}
\begin{tabular}{|c|c|c|c|c|}
\hline
Methods & Area5(mACC) & Area5(mIoU) & 6-fold(mACC) & 6-fold(mIoU) \\
\hline
PointCNN \cite{li2018pointcnn} & 63.9 & 57.3 & 75.6 & 65.4 \\
\hline
PointWeb \cite{zhao2019pointweb} & 66.6 & 60.3 & 76.2 & 66.7 \\
\hline
A-CNN \cite{wu2019pointconv} & - & - & - & 62.9 \\
\hline
DGCNN \cite{wang2019dynamic} & - & - & - & 56.1 \\
\hline
VV-Net \cite{meng2019vv} & - & - & 82.2 & 78.2 \\
\hline
PCCN \cite{wang2018deep} & - & 58.3 & - & - \\
\hline
GAC \cite{wang2019graph} & - & 62.9 & - & - \\
\hline
DPC \cite{engelmann2019dilated} & 68.4 & 61.3 & - & - \\
\hline
SSP+SPG \cite{landrieu2019point} & - & - & 78.3 & 68.4\\
\hline
JSIS3D \cite{pham2019jsis3d} & - & - & 78.6 & - \\
\hline
ASIS \cite{wang2019associatively} & 60.9 & 53.4 & 70.1 & 59.3 \\
\hline
Xu and Lee \cite{xulee2020weakly} & - & 48.0 & - & - \\
\hline
RandLA-Net \cite{Hu_rand_2020_CVPR} & - & - & 82.0 & 70.0 \\
\hline
Tatarchenko et al. \cite{tatarchenko2018tangent} & 62.2 & 52.8 & - & - \\
\hline

\hline
\end{tabular}
\label{table:2}
\end{table*}

\section{Detection, Tracking and Flow Estimation}

\subsection{Overview}
Object detection is a recent research hotspot as the basis of many practical applications. It aims to locate all the objects in the given scene. 3D object detection methods can be generally divided into three categories: multi-view methods, projection-based methods and point-based methods. Figure 4.1 shows an example of 3D object detection on multiple (lidar and camera) views. Aside from image object detection models, the exclusive characteristics of point cloud data provide more potential of optimization. Also, since 3D object tracking and scene flow estimation are two derivative tasks that highly depend on object detection, they will be discussed together in this section.

\begin{figure}[htbp]
\begin{center}
\includegraphics[width=3.2in]{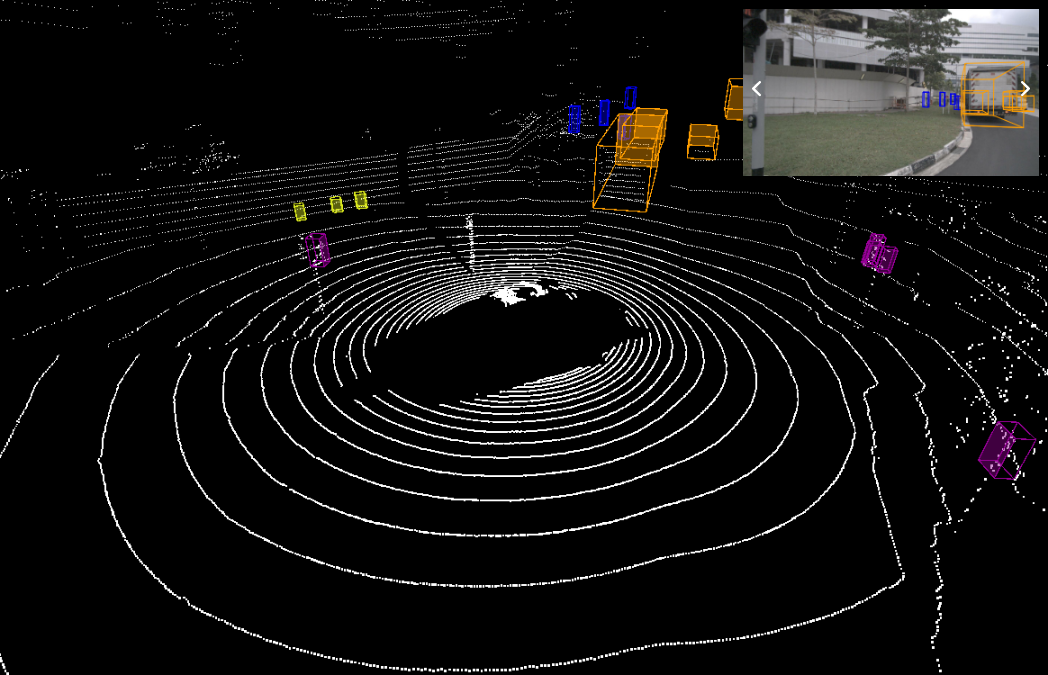}
\end{center}
   \caption{An outdoor scene from nuScenes \cite{caesar2019nuscenes}, annotations in multi-view (lidar/camera) are provided.}
\label{fig:4}
\end{figure}

\subsection{Object Detection}

\subsubsection{Projection-based methods}
The success of convolutional neural networks in image object detection inspired attempts to apply 3D CNN on projected point cloud data. VoxelNet \cite{zhou2018voxelnet} proposed an approach that applies random sampling to the point set within each voxel, and passes them through a novel voxel feature encoding (VFE) layer based on PointNet \cite{qi2017pointnet} and PointNet++ \cite{qi2017pointnet++} to extract point-wise features. A region proposal network is used to produce detection results. Similar to classification models with volumetric representation, VoxelNet runs at a relatively low speed due to the sparsity of voxels and 3D convolutions. SECOND \cite{yan2018second} then proposed an improvement in inference efficiency by taking advantage of sparse convolution network. \par
PointPillars \cite{lang2019pointpillars} utilizes point cloud data in another way. Points are organized in vertical columns (called Pillars), and the features of pillars are extracted with PointNet to generate a pseudo image. The pseudo image is then considered as the input of a 2D object detection pipeline to predict 3D bounding boxes. PointPillars is more accurate than previous fusion approaches, and it is capable of real-time applications with a running speed of 62 FPS. Wang et al. \cite{wang2020} further proposed another anchor-free bounding box prediction based on a cylindrical projection into multi-view features. \par
Projection-based methods suffer from spatial information loss inevitably. Aside from using point-based networks instead, He et al. \cite{He_SAS_2020_CVPR} proposed a structure-aware method to mitigate the problem. The convolutional layers are explicitly supervised to contain structural information by an auxiliary network. The auxiliary network converts the convolutional features from the backbone network to point-level representations and is jointly optimized. After the training process is finished, the auxiliary network can be detached to speed up the inference. \par

\subsubsection{Point-based methods}
Most point-based methods attempt to minimize information loss during feature extraction, and they are the group with the best performance so far. STD \cite{yang2019std} introduced the idea of using sphere anchors for proposal generation, which achieves a high recall with significantly less computation than previous methods. Each proposal is passed through a PointsPool layer that converts proposal features from sparse expression to compact representation, and is robust under transformation. In addition to the regular regression branch, STD has another IoU branch to replace the role of classification score in NMS.\par
Some methods use foreground-background classification to improve the quality of proposals. PointRCNN \cite{shi2019pointrcnn} is such a framework, in which points are directly segmented to screen out foreground points, while semantic features and spatial features are then fused to produce high-quality 3D boxes. Compared with multi-view methods above, segmentation-based methods perform better for complicated scenes and occluded objects.\par
Furthermore, Qi et al. proposed VoteNet \cite{qi2019vote}. A group of points are sampled as seeds, and each seed independently generates a vote for potential center points of objects in the point cloud with the help of PointNet++ \cite{qi2017pointnet++}. By taking advantage of voting, VoteNet outperforms previous approaches on two large indoor benchmarks. However, as the center point prediction of virtual center points is not as stable, the method performs less satisfactorily in wild scenes. As a follow-up work, ImVoteNet \cite{Qi_ImVoteNet_2020_CVPR} inherited the idea of VoteNet and achieved prominent improvement by fusing 3D votes with 2D votes from images.\par
There are also attempts that consider domain knowledge as an auxiliary to enhance features. Associate-3Ddet \cite{Du_A3D_2020_CVPR} introduced the idea of perceptual-to-conceptual association. To enrich perception features that might be incomplete due to occlusion or sparsity, a perceptual-to-conceptual module is proposed to generate class-wise conceptual models from the dataset. The perception and conceptual features will be associated for feature enhancement. \par
Yang et al. \cite{Yang_3DSSD_2020_CVPR} proposed a point-based anchor-free method 3DSSD. This method attempts to reduce computation by abandoning the upsampling layers (e.g. feature propagation layers in \cite{yang2019std}) and refinement stages that are widely used in previous point-based methods. Previous set abstraction layers for downsampling only leverage furthest-point-sampling based on Euclidean distance (D-FPS), instances with a small number of interior points are easily lost under this strategy. In this case, removing upsampling layers could lead to huge performance drop. 3DSSD proposed F-FPS, a new sampling strategy based on feature distances, to preserve more foreground points for instances. The fusion of F-FPS and D-FPS, together with the candidate generation layer and 3D center-ness assignment in the prediction head, help this method outperform previous single-stage methods with a considerable margin. \par
Graph neural networks have also been introduced to 3D object detection for its ability to accommodate intrinsic characteristics of point clouds like sparsity. PointRGCN \cite{zarzar2019pointrgcn} is an early work that introduce graph-based representation for 3D vehicle detection refinement. After that, HGNet \cite{Chen_HGNet_2020_CVPR} introduces a hierarchical graph network based on shape-attentive graph convolution (SA-GConv). By capturing object shapes with relative geometric information and reasoning on proposals, the method obtained a significant improvement on previous results. Besides, Point-GNN \cite{Shi_PointGNN_2020_CVPR} proposed a single-shot method based on graph neural networks. It first builds a fixed radius near-neighbors graph over the input point cloud. Then, the category and the bounding box of affiliation are predicted with the point graph. Finally, a box merging and scoring operation is used to obtain accurate combination of detection results from multiple vertices. \par

\subsubsection{Multi-view methods}
MV3D \cite{chen2017multi} is a pioneering method in multi-view object detection methods on point clouds. In this approach, candidate boxes are generated from BEV map and projected into feature maps of multiple views (RGB images, lidar data, etc.), then the region-wise features extracted from different views are combined to produce the final oriented 3D bounding boxes. While this approach achieves satisfactory performance, much like many other early multi-view methods, its running speed is too slow for practical use. \par
Attempts to improve multi-view methods generally take one of two directions. First, we could find a more efficient way to fuse information from different views. Liang et al. \cite{liang2018deep} use continuous convolutions to effectively fuse feature maps from images and lidar at different resolutions. Image features for each point in BEV space are utilized to generate a dense BEV feature map by bi-linear interpolation with projections of image features within the BEV plane. Experiments show that dense BEV feature maps perform better than discrete image feature maps and sparse point cloud feature maps. Second, many methods propose innovative feature extraction approaches to obtain representations of input data with higher robustness. SCANet \cite{lu2019scanet} introduced a Spatial Channel Attention (SCA) module to make use of multi-scale contextual information. The SCA module captures useful features from the global and multi-scale context of given scene, while an Extension Spatial Unsample (ESU) module helps combine multi-scale low-level features to generate high-level features with rich spatial information, which then leads to accurate 3D object proposals. In RT3D \cite{zeng2018rt3d}, the majority of convolution operations prior to the RoI pooling module are removed. With such optimization, RoI convolutions only need to be performed once for all proposals, accelerating the method to run at 11.1 FPS, which is five times faster than MV3D \cite{chen2017multi}. \par
Another approach to detect 3D objects is to generate candidate regions on 2D plane with 2D object detectors, then extract a 3D frustum proposal for each 2D candidate region. In F-PointNets \cite{qi2018frustum}, each 2D region generates a frustum proposal, and the features of each 3D frustum are learned with PointNet \cite{qi2017pointnet} or PointNet++ \cite{qi2017pointnet++} and used for 3D bounding box estimation. PointFusion \cite{xu2018pointfusion} uses both 2D image region and corresponding frustum points for more accurate 3D box regression. A fusion network is proposed to directly predict corner locations of boxes by fusing image features and global features from point clouds. \par

\subsection{Object Tracking}
Object tracking targets estimating the location of a certain object in subsequent frames given its state in the first frame. The success of Siamese networks \cite{bertinetto2016fully} in 2D image object tracking inspired 3D object tracking, and Giancola et al. \cite{giancola2019leveraging} extend Siamese networks to 3D. In this method, candidates are first generated by a Kalman filter, then passed through an encoding model to generate compact representations with shape regularization, and match the detected objects by cosine similarity. Zarzar et al. \cite{zarzar2019pointrgcn} proposed another method that captures target objects more efficiently by leveraging a 2D Siamese network to detect coarse object candidates on BEV representation. The coarse candidates are then refined by cosine similarity in the 3D Siamese network.\par
Chiu et al. \cite{chiu2020probabilistic} introduced the Kalman filter to encode the hidden states of objects. The state of an object is represented by a tuple of 11 variables, including position, orientation, size and speed. A Kalman filter is adopted to predict the object in next frame based on previous information, and a greedy algorithm is used for data association with Mahalanobis distance.\par
Besides, Qi et al. \cite{Qi_P2B_2020_CVPR} proposed P2B, a point-to-box method for 3D object tracking. It divides the task into two parts. The first part is target-specific feature augmentation, seeds from the template and the search area are generated with a PointNet++ backbone, and the search area seeds will be enriched with target clues from the template. The second is target proposal and verification, candidate target centers are regressed and seed-wise targetness is evaluated for joint target proposal and verification. \par
\subsection{Scene Flow Estimation}
Similar to optical flow estimation on images, 3D scene flow estimation works on a sequence of point clouds. FlowNet3D \cite{liu2019flownet3d} is a representative work that directly estimates scene flows from pairs of consecutive point clouds. The flow embedding layer is used to learn point-level features and motion features. The experiment results of FlowNet3D shows that it performs less than satisfactorily in non-static scenes, and the angles of predicted motion vectors sometimes significantly differ from the ground truth. FlowNet3D++ \cite{wang2020flownet3d++} is proposed to fix these issues by introducing a cosine distance loss in angles, and a point-to-plane distance loss to improve accuracy in dynamic scenes. HPLFlowNet \cite{gu2019hplflownet}, on the other hand, proposed a series of bilateral convolutional layers to fuse information from two consecutive frames and restore structural information from unconstructed point clouds. \par
In addition, MeteorNet \cite{liu2019meteornet} introduced direct grouping and chained-flow grouping to group temporal neighbors, and adopted information aggregation over neighbor points to generate representation for dynamic scenes. Derived from recurrent models in images, Fan and Yang \cite{fan2019pointrnn} proposed PointRNN, PointGRU and PointLSTM to encode dynamic point clouds by capturing both spatial and temporary information.

\subsection{Experiments}
KITTI \cite{Geiger2013IJRR,Geiger2012CVPR,Fritsch2013ITSC,Menze2015CVPR} is one of the most popular benchmarks for many computer vision tasks, including those in images, point clouds, and multi-views. By taking advantage of autonomous driving platforms, KITTI provides raw data of real-world scenes, and allows evaluation on multiple tasks. Table \ref{table:3} shows experimental results of different methods on KITTI. Some methods, such as VoteNet \cite{ding2019votenet}, which does not provide detailed test results on KITTI, are not listed.

\begin{table*}[htbp]
\centering
\caption{Experiment results on KITTI 3D detection benchmark, E/M/H stands for easy/medium/hard samples.}
\begin{tabular}{|c|c|c|c|c|c|c|c|c|c|c|c|}
\hline
\multirow{2}*{Method} & \multirow{2}*{Category} & \multirow{2}*{Speed} & \multicolumn{3}{|c|}{Car} & \multicolumn{3}{|c|}{Pedestrians} & \multicolumn{3}{|c|}{Cyclists} \\
\cline{4-12}
~ & ~ & ~ & E & M & H & E & M & H & E & M & H\\
\hline
MV3D \cite{chen2017multi} & multi-view & 2.8 & 74.8 & 63.6 & 54.0 & - & - & - & - & - & - \\
\hline
AVOD \cite{ku2018joint} & multi-view & 12.5 & 89.8 & 85.0 & 78.3 & 42.6 & 33.6 & 30.1 & 64.1 & 48.1 & 42.4 \\
\hline
SCANet \cite{lu2019scanet} & multi-view & 12.5 & 76.4 & 66.5 & 60.2 & - & - & - & - & - & - \\
\hline
PIXOR \cite{yang2018pixor} & projection & 28.6 & 84.0 & 80.0 & 74.3 & - & - & - & - & - & - \\
\hline
VoxelNet \cite{zhou2018voxelnet} & projection & 2.0 & 77.5 & 65.1 & 57.7 & 39.5 & 33.7 & 31.5 & 61.2 & 48.4 & 44.4 \\
\hline
SECOND \cite{yan2018second} & projection & 26.3 & 83.3 & 72.6 & 65.8 & 49.0 & 38.8 & 34.9 & 71.3 & 52.1 & 45.8 \\
\hline
PointPillars \cite{lang2019pointpillars} & projection & 62.0 & 82.6 & 74.3 & 69.0 & 54.5 & 41.2 & 38.9 & 77.1 & 85.7 & 52.0 \\
\hline
PointRCNN \cite{shi2019pointrcnn} & point & 10.0 & 87.0 & 75.6 & 70.7 & 48.0 & 39.4 & 36.0 & 75.0 & 58.8 & 52.5 \\
\hline
PointRGCN \cite{zarzar2019pointrgcn} & point & 3.8 & 86.0 & 95.6 & 70.7 & - & - & - & - & - & - \\
\hline
STD \cite{yang2019std} & point & 12.5 & 88.0 & 79.7 & 75.1 & 53.3 & 42.5 & 38.3 & 78.7 & 61.6 & 55.3 \\
\hline
Point-GNN \cite{Shi_PointGNN_2020_CVPR} & point & - & 88.3 & 79.5 & 72.3 & 52.0 & 43.8 & 40.1 & 78.6 & 63.5 & 57.0 \\
\hline
PV-RCNN \cite{Shi_PVRCNN_2020_CVPR} & point & - & 90.2 & 81.4 & 76.8 & 52.1 & 43.3 & 40.3 & 78.6 & 63.7 & 57.7 \\
\hline
3DSSD \cite{Yang_3DSSD_2020_CVPR} & point & 26.3 & 88.4 & 79.6 & 74.6 & 54.6 & 44.3 & 40.2 & 82.5 & 64.1 & 56.9 \\
\hline
\end{tabular}
\label{table:3}
\end{table*}

\section{Registration}

\subsection{Overview}
In some scenarios like autopilot, it is of great value to find the relationship between point cloud data of the same scene collected in different ways. These data might be collected from different angles, or at different times. 3D point cloud registration (sometimes also called matching) attempts to align two or more different point clouds by estimating the transformation between them. It is a challenging problem affected by a lot of factors including noise, outliers and nonrigid spatial transformation.

\subsection{Traditional Methods}
The Iterative Closest Point (ICP) algorithm \cite{besl1992method} is a pioneering work that solves 3D point set registration. The basic pipeline of ICP and its variants is as follows: (1) Sample a point set $P$ from the source point cloud. (2) Compute the closest point set $Q$ from the target point cloud. (3) Calculate the registration (transformation) with $P$ and $Q$. (4) Apply the registration, and if the error is above some threshold, go back to step (2), otherwise terminate. A global refinement step is usually required for better performance. The performance of ICP highly depends on the quality of initialization and whether the input point clouds are clean. Generalized-ICP \cite{segal2009generalized} and Go-ICP \cite{yang2015go} are two representative follow-up works that mitigate the problems of ICP in different ways. \par
Coherent Point Drift (CPD) algorithm \cite{myronenko2010point} considers the alignment as a problem of probability density estimation. Concretely, the algorithm consider the first point set as the Gaussian mixture model centroids, and the transformation is estimated by maximizing the likelihood in fitting them to the second point set. The movement of these centroids are forced to be coherent to preserve the topological structure. \par
Robust Point Matching (RPM) \cite{gold1998new} is another influential point matching algorithm. The algorithm starts with soft assignments of the point correspondences, and these soft assignments will get hardened through deterministic annealing. RPM is generally more robust than ICP, but still sensitive to initialization and noise.\par
Iglesias et al. \cite{Iglesias_2020_CVPR} focused on the registration of several point clouds to a global coordinate system. In other words, with the original set of $n$ points, we want to find the correspondences between (subsets of) the original set and $m$ local coordinate systems respectively. Iglesias et al. consider the problem as a Semidefinite Program (SDP), and attempt to analyze it with the application of Lagrangian duality. \par

\subsection{Learning-based Methods}
DeepVCP \cite{lu2019deepvcp} is the first end-to-end learning-based framework in point cloud registration. Given the source and target point cloud, PointNet++ \cite{qi2017pointnet++} is applied to extract local features. A point weighting layer then helps select a set of $N$ keypoints, after which $N\times C$ candidates from the target point cloud are selected and passed through a deep feature embedding operation together with keypoints from the source. Finally, a corresponding point generation layer takes the embeddings and generates the final result. Two losses are incurred: (1) the Euclidean distance between  the estimated corresponding points and ground truth under the ground truth transformation, and (2) the distance between the target under the estimated transformation and ground truth. These losses are combined to consider both global geometric information and local similarity. \par
3DSmoothNet \cite{gojcic2019perfect} is proposed to perform 3D point cloud matching with a compact learned local feature descriptor. Given two raw point clouds as input, the model first computes the local reference frame (LRF) of the neighborhood around the randomly sampled interest points. Then the neighborhoods are transformed into canonical representations and voxelized by Gaussian smoothing, and the local feature of each point is then generated by 3DSmoothNet. The features will then be utilized by a RANSAC approach to produce registration results. The proposed smooth density value (SDV) voxelization outperforms traditional binary-occupancy grids by reducing the impact of boundary effects and noise, and provides greater compactness. Following 3DSmoothNet, Gojcic et al. \cite{gojcic2020learning} proposed another method that formulates conventional two-stage approaches in an end-to-end structure. Earlier methods solve the problem in two steps, the pairwise alignment and the globally consistent refinement, by jointly learning both parts. Gojcic et al.'s method outperforms previous ones with higher accuracy and less computational complexity. \par
RPM-Net \cite{yew2020rpm} inherits the idea of RPM \cite{gold1998new} algorithm, and takes advantage of deep learning to enhance robustness against noise, outliers and bad initialization. In this method, the initialization assignments are generated based on hybrid features from a network instead of spatial distances between points. The parameters of annealing is predicted by a secondary network, and a modified Chamfer distance is introduced to evaluate the quality of registration. This method outperforms previous methods no matter the input is clean, noisy, or even partially visible.

\section{Augmentation and Completion}

\subsection{Overview}
Point clouds collected by lidar, especially those from outdoor scenes, suffer from different kinds of quality issues like noise, outliers, and missing points. Many attempts have been made to improve the quality of raw point clouds by completing missing points, removing outliers and so on. The motivation and implementation vary a lot among different approaches; in this paper, we divide them into two categories: discriminative models and generative models.

\subsection{Discriminative Methods}
Noise in point clouds collected from outdoor scenes is naturally inevitable. To prevent noise from influencing the encoding of point clouds, some denoising methods shall be applied in pre-processing. Conventional methods include local surface fitting, neighborhood averaging and guessing the underlying noise model. PointCleanNet \cite{rakotosaona2020pointcleannet} proposed a data-driven method to remove outliers and reduce noise. With a deep neural network adapted from PCPNet\cite{guerrero2018pcpnet}, the model first classifies outliers and discards them, then estimates a correction projection that projects noise to original surfaces.\par
Hermosilla et al. \cite{hermosilla2019total} proposed Total Denoising that achieved unsupervised denoising of 3D point clouds without additional data. The unsupervised image denoisers are usually built based on the assumption that the value of a noisy pixel follows a distribution around a clean pixel value. Under this assumption, the original clean value can be recovered by learning the parameters of the random distribution. However, such an idea cannot be directly extended to point clouds because there are multiple formats of noise in point clouds, such as a global position deviation where no reliable reference point exists. Total Denoising introduces a spatial prior term that finds the closest of all possible modes on a manifold. The model achieves competitive performance against supervised models.\par
While a lot of models benefit from rich information in dense point clouds, some others are suffering from the low efficiency with large amounts of points. Conventional downsampling approaches usually have to risk dropping critical points. Nezhadarya et al. \cite{Nezhadarya_2020_CVPR} proposed the critical points layer (CPL) that learns to reduce the number of points while preserving the important ones. The layer is deterministic, order-agnostic and also efficient by avoiding neighbor search. Aside from that, SampleNet \cite{Lang_sample_2020_CVPR} proposed a differentiable relaxation of point sampling by approximating points after sampling as a mixture of original points. The method has been tested as a front to networks on various tasks, and obtains decent performance with only a small fraction of the raw input point cloud. \par

\subsection{Generative Methods}
Generative adversarial networks are widely studied for 2D images and CNNs, as they help locate the potential defects of networks by generating false samples. While typical applications of point cloud models, such as autonomous driving, consider safety as a critical concern, it is helpful to study how current deep neural networks on point clouds are affected by false samples.\par
Xiang et al. \cite{xiang2019generating} proposed several algorithms to generate adversarial point clouds against PointNet. The adversarial algorithms work in two ways: point perturbation and point generation. Perturbation is implemented by shifting existing points negligibly, and generation is implemented by either adding some independent and scattered points or a small number of point clusters with predefined shapes. Shu et al. \cite{shu20193d} proposed tree-GAN, a tree-structured graph convolution network. By performing graph convolution within a tree, the model takes advantage of ancestor information to enrich the capacity of features. Along with the development of adversarial networks, DUP-Net \cite{zhou2019dup} is proposed to defend 3D adversarial models. The model contains a statistical outlier removal (SOR) module as denoiser and a data-driven upsampling network as upsampler. \par
Aside from adversarial generation, generative models are also used for point cloud upsampling. There are generally two motivations to upsample a point cloud. The first is to reduce the sparseness and irregularity of data, and the second is to restore missing points due to occlusion. \par
For the first aim, PU-Net \cite{yu2018pu} proposed upsampling in the feature space. For each point, multi-level features are extracted and expanded via a multi-branch convolution unit; after that, the expanded feature is split into multiple features and reconstructed to upsample the input set. Inspired by image super-resolution models, Wang et al. \cite{yifan2019patch} proposed a cascade of patch-based upsampling networks, learning different levels of details at different steps, where at each step the network focuses only on a local patch from the output of the previous step. The architecture is able to upsample a sparse input point set to a dense set with rich details. Hui et al. \cite{hui2020progressive} also proposed a learning-based deconvolution network that generates multi-resolution point clouds based on low-resolution input with bilateral interpolation performed in both the spatial and feature spaces.\par
Meanwhile, early methods in completion, such as \cite{dai2017shape}, tend to voxelize the input point cloud at the very beginning. PCN \cite{yuan2018pcn} was the first framework to work on raw point clouds and in a coarse-to-fine fashion. Wang et al. \cite{Wang_Comp_2020_CVPR} improved the results with a two-step reconstruction design. Besides, Huang et al. \cite{Huang_PF_2020_CVPR} proposed PF-Net that preserves the spatial structure of the original incomplete point cloud, and predicts the missing points hierarchically a multi-scale generating network. GRNet\cite{xie2020grnet}, on the other hand, proposed a gridding-based which retrieve structural context by performing cubic feature sampling per grid, and complete the output with "Gridding Reverse" layers and MLPs. \par
Lan et al. \cite{lan2019robust} proposed a probabilistic approach to optimize outliers by applying EM algorithm with Cauchy-Uniform mixture model to suppress potential outliers. More generally, PU-GAN \cite{li2019pu} proposed a data-driven generative adversarial network to learn point distributions from the data and upsample points over patches on the surfaces of objects. Furthermore, RL-GAN-Net \cite{sarmad2019rl} uses a reinforcement learning (RL) agent to provide fast and reliable control of a generative adversarial network. By first training the GAN on the dimension-reduced latent space representation, and then finding the correct input to generate the representation that fits the current input form the uncompleted point cloud with a RL agent, the framework is able to convert noisy, partial point cloud into a completed shape in real time.\par

\section{Conclusion}
In this paper, we reviewed milestones and recent progress on various problems in 3D point clouds. With the expectation of practical applications like autonomous driving, point cloud understanding has received increasing attention lately. In 3D shape classification, point-based models have achieved satisfactory performance on recognized benchmarks. Methods developed from image tasks, such as two-stage detector and the Siamese architecture, are widely introduced in 3D segmentation, object detection and other derivative tasks. Specific deep learning frameworks are proposed to match point clouds of the same scene from multiple scans, and generative networks are adapted to improve the quality of point cloud data with noise and missing points. Deep learning methods with proper adaption have been proven to efficiently help overcome the unique challenges in point cloud data. \par

{\small
\bibliographystyle{ieee_fullname}
\bibliography{final}
}

\end{document}